\documentclass{article}

\usepackage[accepted]{icml2008}

\usepackage[T1]{fontenc}

\usepackage[centertags]{amsmath}
\usepackage{amsxtra}
\usepackage{amstext}
\usepackage{amsfonts}
\usepackage{amssymb}

\usepackage{mathrsfs}
\usepackage{dsfont}

\usepackage[round]{natbib}
\makeatletter
\let\mlapa@natlbibitem\@lbibitem
\def\@lbibitem[#1]{\@ifnextchar[{\mlapa@lbibitem{#1}}{\mlapa@natlbibitem[#1]}}
\def\mlapa@lbibitem#1[#2][#3]#4{\mlapa@split#1\@nil{#3}{#4}}
\def\mlapa@split#1, #2\@nil#3#4{\mlapa@natlbibitem[#1(#3)]{#4}}
\makeatother
\usepackage{datetime}

\newtheorem{theorem}{Theorem}

\newtheorem{lemma}[theorem]{Lemma}

\newenvironment{proof}[1][]{%
  \noindent\textsc{Proof #1~}%
  }{%
  \hspace*{\fill}~\mbox{$\Box$}%
  \vspace{2ex}
  }

\newcommand{\A}{\mathcal{A}}

\newcommand{\E}{\mathbb{E}}
\newcommand{\EE}[2][]{\mathbb{E}_{#1}\left[#2\right]}

\renewcommand{\H}{\mathbf{H}}
\newcommand{\I}{\mathbb{I}}

\newcommand{\M}{\mathcal{M}}

\newcommand{\PP}[2][]{\mathbb{P}_{#1}\left[#2\right]}
\renewcommand{\P}{\mathsf{P}}
\newcommand{\Q}{\mathcal{Q}}
\newcommand{\R}{\mathds{R}}

\newcommand{\X}{\mathcal{X}}

\newcommand{\eps}{\varepsilon}

\newcommand{\dprod}[1]{\left<#1\right>}
\newcommand{\norm}[1]{\left\|#1\right\|}
\newcommand{\snorm}[1]{\|#1\|}
\newcommand{\abs}[1]{\left|#1\right|}
\newcommand{\set}[1]{\left\{#1\right\}}
\newcommand{\sset}[1]{\{#1\}}

\newcommand{\argmax}[1]{\underset{#1}{\arg\max}\:}

\newcommand{\MDP}[1][]{(\X_{#1},\A_{#1},\P_{#1},r_{#1},\gamma)}
\newcommand{\eg}{\textit{e.g.},~}

\newcommand{\ie}{\textit{i.e.},~}

\newcommand{\SARSA}{\textsf{SARSA}}

\icmltitlerunning{An Analysis of Reinforcement Learning with Function Approximation}

\begin{document} 

\twocolumn[

\icmltitle{An Analysis of Reinforcement Learning with Function Approximation}

\icmlauthor{Francisco S. Melo}{fmelo@cs.cmu.edu}
\icmladdress{Carnegie Mellon University,
             Pittsburgh, PA 15213, USA}
\icmlauthor{Sean P. Meyn}{meyn@control.csl.uiuc.edu}
\icmladdress{Coordinated Science Lab,
						Urbana, IL 61801, USA}
\icmlauthor{M. Isabel Ribeiro}{mir@isr.ist.utl.pt}
\icmladdress{Institute for Systems and Robotics,
             1049-001 Lisboa, Portugal}

\vskip 0.3in
]

% \twocolumn[...] typesets its argument inside a group, so recent versions of fancyhdr
% lose the running head set by \icmltitle; set it again here.
\makeatletter\chead{\small\bf\@icmltitlerunning}\makeatother

\begin{abstract}

We address the problem of computing the optimal $Q$-function in Markov decision problems with infinite state-space. We analyze the convergence properties of several variations of $Q$-learning when combined with function approximation, extending the analysis of TD-learning in \citep{Vanroy96-b} to stochastic control settings. We identify conditions under which such approximate methods converge with probability 1. We conclude with a brief discussion on the general applicability of our results and compare them with several related works.

\end{abstract}

\section{Introduction}%
\label{Sec:Intro}

Convergence of $Q$-learning with function approximation has been a long standing open question in reinforcement learning \citep{Sutton99}. In general, value-based reinforcement learning (RL) methods for optimal control behave poorly when combined with function approximation \citep{Baird95icml,Vanroy96-b}. In this paper, we address this problem by analyzing the convergence of $Q$-learning when combined with linear function approximation. We identify a set of conditions that imply the convergence of this approximation method with probability 1 (w.p.1), when a fixed learning policy is used, and provide an interpretation of the resulting approximation as the fixed point of a Bellman-like operator. This motivates the analysis of several variations of $Q$-learning when combined with linear function approximation. In particular, we study a variation of $Q$-learning using importance sampling and an on-policy variant of $Q$-learning (\SARSA).

The paper is organized as follows. We start in Section~\ref{Sec:MDPs} by describing Markov decision problems. We proceed with our analysis of the $Q$-learning algorithm and its variants, and produce our main results in Section~\ref{Sec:RLFA}. We also compare our results with other related works in the RL literature. We conclude with some further discussion in Section~\ref{Sec:Conc}.

\section{Markov Decision Problems}%
\label{Sec:MDPs}

Let $\MDP$ be a Markov decision problem (MDP) with a compact state-space $\X\subset\R^p$ and a finite action set $\A$. The action-dependent kernel $\P_a$ defines the transition probabilities for the underlying controlled Markov chain $\set{X_t}$ as
\begin{displaymath}
  \PP{X_{t+1}\in U\mid{X}_t=x,A_t=a}=\P_a(x,U),
\end{displaymath}
where $U$ is any measurable subset of $\X$. The $\A$-valued process $\set{A_t}$ represents the control process: $A_t$ is the control action at time instant $t$.%
\footnote{We take the control process $\set{A_t}$ to be adapted to the $\sigma$-algebra induced by $\set{X_t}$.} 
Solving the MDP consists in determining the control process $\set{A_t}$ maximizing the expected total discounted reward 
\begin{displaymath}
  V\big(\set{A_t},x\big)=\EE{\sum_{t=0}^\infty\gamma^tR(X_t,A_t)\mid X_0=x},
\end{displaymath}
where $0\leq\gamma<1$ is a discount-factor and $R(x,a)$ represents a random ``reward'' received for taking action $a\in\A$ in state $x\in\X$. For simplicity of notation, we consider a bounded deterministic function $r:\X\times\A\times\X\longrightarrow\R$ assigning a reward $r(x,a,y)$ every time a transition from $x$ to $y$ occurs after taking action $a$. This means that
\begin{displaymath}
  \EE{R(x,a)}=\int_\X r(x,a,y)\P_a(x,dy).
\end{displaymath}
The \emph{optimal value function} $V^*$ is defined for each state $x\in\X$ as 
\begin{align*}
  V^*(x)&=\max_{\set{A_t}}V\big(\set{A_t},x\big)=\\
	  	  &=\max_{\set{A_t}}\EE{\sum_{t=0}^\infty\gamma^tR(X_t,A_t)\mid{X_0}=x}
\end{align*}
and verifies the Bellman optimality equation
\begin{equation}\label{Eq:Bellman}
  V^*(x)=\max_{a\in\A}\int_\X\big(r(x,a,y)+\gamma{V}^*(y)\big)\P_a(x,dy).
\end{equation}
$V^*(x)$ represents the expected total discounted reward received along an optimal trajectory starting at state $x$. We can also define the optimal $Q$-function $Q^*$ as
\begin{equation}\label{Eq:Qfunction}
  Q^*(x,a)=\int_\X\big(r(x,a,y)+\gamma{}V^*(y)\big)\P_a(x,dy),
\end{equation}
representing the expected total discounted reward along a trajectory starting at state $x$ obtained by choosing $a$ as the first action and following the optimal policy thereafter. The control process $\set{A_t}$ defined as
\begin{displaymath}
  A_t=\argmax{a\in\A}Q^*(X_t,a),\quad\forall{t},
\end{displaymath}
is optimal in the sense that $V(\set{A_t},x)=V^*(x)$ and defines a mapping $\pi^*:\X\to\A$ known as the \emph{optimal policy}. The optimal policy determines the optimal decision rule for a given MDP.

More generally, a (Markov) \emph{policy} is any mapping $\pi_t$ defined over $\X\times\A$ generating a control process $\set{A_t}$ verifying, for all $t$,
\begin{displaymath}
  \PP{A_t=a\mid X_t=x}=\pi_t(x,a),\quad\forall{t}.
\end{displaymath}
We write $V^{\pi_t}(x)$ instead of $V(\set{A_t},x)$ if the control process $\set{A_t}$ is generated by policy $\pi_t$. A policy $\pi_t$ is \emph{stationary} if it does not depend on $t$ and \emph{deterministic} if it assigns probability $1$ to a single action in each state and is thus represented as a map $\pi_t:\X\to\A$ for every $t$. Notice that the optimal control process can be obtained from the optimal (stationary, deterministic) policy $\pi^*$, which can in turn be obtained from $Q^*$. Therefore, the optimal control problem is solved once the function $Q^*$ is known for all pairs $(x,a)$.

Now given any real function $q$ defined over $\X\times\A$, we define the Bellman operator
\begin{equation}\label{Eq:BellmanOperator}
  (\mathbf{H}q)(x,a)=\int_\X\big(r(x,a,y)+\gamma\max_{u\in\A}q(y,u)\big)\P_a(x,dy).
\end{equation}
The function $Q^*$ in \eqref{Eq:Qfunction} is the fixed-point of $\mathbf{H}$ and, since this operator is a contraction in the sup-norm, a fixed-point iteration can be used to determine $Q^*$ (at least theoretically).

\subsection{The $Q$-Learning Algorithm}%
\label{Subsec:QL}

We previously suggested that a fixed-point iteration could be used to determine the function $Q^*$. In practice, this requires two important conditions:
\begin{itemize}
  \item The kernel $\P$ and the reward function $r$ are known;
  \item The successive estimates for $Q^*$ can be represented compactly and stored in a computer with finite memory.
\end{itemize}
If $\P$ and/or $r$ are not known, a fixed-point iteration using $\mathbf{H}$ is not possible. To solve this problem, Watkins proposed in 1989 the \emph{$Q$-learning algorithm} \citep{Watkins89phd}. $Q$-learning proceeds as follows: consider a MDP $\M=\MDP$ and suppose that $\set{x_t}$ is an infinite sample trajectory of the underlying Markov chain obtained with some policy $\pi_t$. The corresponding sample control process is denoted as $\set{a_t}$ and the sequence of obtained rewards as $\set{r_t}$. Given any initial estimate $Q_0$, $Q$-learning successively updates this estimate using the rule
\begin{equation}\label{Eq:QL}
  Q_{t+1}(x,a)=Q_t(x,a)+\alpha_t(x,a)\Delta_t,
\end{equation}
where $\set{\alpha_t}$ is a step-size sequence and $\Delta_t$ is the temporal difference at time $t$,
\begin{equation}\label{Eq:TD}
  \Delta_t=r_t+\gamma\max_{b\in\A}Q_t(x_{t+1},b)-Q_t(x_t,a_t).
\end{equation}
If both $\X$ and $\A$ are finite sets, each estimate $Q_t$ is simply a $\abs{\X}\times\abs{\A}$ matrix and can be represented explicitly in a computer. In that case, the convergence of $Q$-learning and several other related algorithms (such as TD$(\lambda)$ or \SARSA) has been thoroughly studied (see, for example, \citep{Bertsekas96} and references therein). However, if either $\X$ or $\A$ are infinite or very large, explicitly representing each $Q_t$ becomes infeasible and some form of compact representation is needed (\eg using function approximation). In this paper, we address how several RL methods such as $Q$-learning and {\SARSA} can be combined with function approximation and still retain their main convergence properties.

\section{Reinforcement Learning with Linear Function Approximation}%
\label{Sec:RLFA}

In this section, we address the problem of determining the optimal $Q$-function for MDPs with infinite state-space $\X$. Let $\Q=\set{Q_\theta}$ be a family of real-valued functions defined in $\X\times\A$. It is assumed that the function class is linearly parameterized, so that  $\Q$ can be expressed as the linear span of a \emph{fixed set} of $M$ functions $\phi_i:\X\times\A\to\R$. For each $M$-dimensional parameter vector $\theta\in\R^M$, the function $Q_\theta\in \Q$ is defined as,
\begin{displaymath}
  Q_\theta(x,a)=\sum_{i=1}^M\phi_i(x,a)\theta(i)=\phi^\top(x,a)\theta,
\end{displaymath}
where $^\top$ represents the transpose operator. We will also denote the above function by $Q(\theta)$ to emphasize the dependence on $\theta$ over the dependency on $(x,a)$.

Let $\pi$ be a fixed stochastic, stationary policy and suppose that $\set{x_t}$, $\set{a_t}$ and $\set{r_t}$ are sampled trajectories of states, actions and rewards obtained from the MDP using policy $\pi$. In the original $Q$-learning algorithm, the $Q$-values are updated according to \eqref{Eq:QL}. The temporal difference $\Delta_t$ can be interpreted as a $1$-step estimation error with respect to the optimal function $Q^*$. The update rule in $Q$-learning ``moves'' the estimates $Q_t$ closer to the desired function $Q^*$, minimizing the expected value of $\Delta_t$.

In our approximate setting, we apply the same underlying idea to obtain the update rule for approximate $Q$-learning:
\begin{equation}\label{Eq:QLFA}
  \begin{split}
    \theta_{t+1}&=\theta_t+\alpha_t\nabla_{\theta}Q_\theta(x_t,a_t)\Delta_t\\
  	  			&=\theta_t+\alpha_t\phi(x_t,a_t)\Delta_t
  \end{split}
\end{equation}
where, as above, $\Delta_t$ is the temporal difference at time $t$ defined in \eqref{Eq:TD}. Notice that \eqref{Eq:QLFA} updates $\theta_t$ using the temporal difference $\Delta_t$ as the error. The gradient $\nabla_{\theta}Q_\theta$ provides the ``direction'' in which this update is performed.

To establish convergence of the algorithm \eqref{Eq:QLFA} we adopt an ODE argument, establishing the trajectories of the algorithm to closely follow those of an associated ODE with a globally asymptotically stable equilibrium point. This will require several regularity properties on the policy $\pi$ and on its induced Markov chain that will, in turn, motivate the study of the on-policy version of the algorithm. This on-policy algorithm can be seen as an extension of {\SARSA} to infinite settings.

\subsection{Convergence of $Q$-Learning}%
\label{Subsec:Thms}

We now proceed by identifying conditions that ensure the convergence of $Q$-learning with linear function approximation as described by \eqref{Eq:QLFA}. Due to space limitations, we overlook some of the technical details in the proofs that can easily be filled in.

We start by introducing some notation that will greatly simplify the presentation. Given an MDP $\M=\MDP$ with compact state space $\X\subset\R^p$, let $(\X,\P_\pi)$ be the Markov chain induced by a fixed policy $\pi$. We assume the chain $(\X,\P_\pi)$ to be uniformly ergodic with invariant probability measure $\mu_X$ and the policy $\pi$ to verify $\pi(x,a)>0$ for all $a\in\A$ and $\mu_X$-almost all $x\in\X$. We denote by $\mu_\pi$ the probability measure defined for each measurable set $U\subset\X$ and each action $a\in\A$ as
\begin{displaymath}
  \mu_\pi(U\times\set{a})=\int_U\pi(x,a)\mu_X(dx).
\end{displaymath}
Let now $\set{\phi_i,i=1,\ldots,M}$ be a set of basis functions, which we assume bounded and linearly independent in $L^2(\mu_\pi)$, to be used in our approximate $Q$-learning algorithm. We denote by $\Sigma_\pi$ the matrix defined as
\begin{align*}
  \Sigma_\pi
    &=\EE[\pi]{\phi(x,a)\phi^\top(x,a)}=\int_{\X\times\A}\phi\:\phi^\top d\mu_\pi
\end{align*}
Notice that the above expression is well-defined and independent of the initial distribution for the chain, due to our assumption of uniform ergodicity. 

For fixed $\theta\in\R^M$ and $x\in\X$, define the set of maximizing actions at $x$ as
\begin{displaymath}
  \A^\theta_x=\sset{a^*\in\A\mid\phi^\top(x,a^*)\theta=\max_{a}\phi^\top(x,a)\theta}
\end{displaymath}
and the greedy policy with respect to $\theta$ as any policy that, at each state $x$, assigns positive probability only to actions in $\A_x^\theta$. Finally, let $\phi_x$ denote the row-vector $\phi^\top(x,a)$, where $a$ is a random action generated according to the policy $\pi$ at state $x$; likewise, let $\phi_x^\theta$ denote the row-vector $\phi^\top(x,a_x^\theta)$, where $a_x^\theta$ is now any action in $\A_x^\theta$. We now introduce the \emph{$\theta$-dependent} matrix
\begin{displaymath}
  \Sigma^*_\pi(\theta)=\EE[\pi]{\big(\phi_x^\theta\big)^\top\phi_x^\theta}.
\end{displaymath}
By construction, $\Sigma_\pi$ is positive definite and each $\Sigma^*_\pi(\theta)$ is positive semidefinite. Notice also the difference between $\Sigma_\pi$ and each $\Sigma^*_\pi$: the actions in the definition of the former are taken according to $\pi$ while in the latter they are taken greedily with respect to a particular $\theta$.

We are now in position to introduce our first result.

\begin{theorem}\label{Theo:QLFAConv}
Let $\M$, $\pi$ and $\set{\phi_i,i=1,\ldots,M}$ be as defined above. If there is $\eta>0$ such that, for all $\theta$,
\begin{equation}\label{Eq:Cond01}
  \Sigma_\pi\geq2(1+\eta)\gamma^2\Sigma^*_\pi(\theta)
\end{equation}
and the step-size sequence verifies
\begin{align*}
  \sum_t\alpha_t&=\infty & \sum_t\alpha_t^2&<\infty,
\end{align*}
then the algorithm in \eqref{Eq:QLFA} converges w.p.1 and the limit point $\theta^*$ verifies the recursive relation
\begin{displaymath}
  Q(\theta^*)=\Pi_\Q\H{Q}(\theta^*),
\end{displaymath}
where $\Pi_\Q$ is the orthogonal projection onto $\Q$.%
\footnote{
The orthogonal projection is naturally defined in the (infinite-dimensional) Hilbert space containing $\Q$ with inner-product given by
\begin{displaymath}
  \dprod{f,g}=\int_{\X\times\A}f\:g\:d\mu_\pi.
\end{displaymath}
}
\end{theorem}
\begin{proof}
We establish the main statement of the theorem using a standard ODE argument.

The assumptions on the chain $(\X,\P_\pi)$ and basis functions $\set{\phi_i,i=1,\ldots,M}$ and the fact that $\pi(x,a)>0$ for all $a\in\A$ and $\mu_X$-almost every $x\in\X$ ensure the applicability of Theorem~17 in page~239 of \citep{Benveniste90}. Therefore, the convergence of the algorithm can be analyzed in terms of the stability of the equilibrium points of the associated ODE
\begin{equation}\label{Eq:ODE01}
    \dot{\theta}=\E_{\pi}\Big[\phi_x^\top\big(r(x,a,y)+\gamma\phi^\theta_y\theta-\phi_x\theta\big)\Big],
\end{equation}
where we omitted the explicit dependence of $\theta$ on $t$ to avoid excessively cluttering the expression. If the ODE \eqref{Eq:ODE01} has a globally asymptotically stable equilibrium point, this implies the algorithm \eqref{Eq:QLFA} to converge w.p.1 \citep{Benveniste90}. Let then $\theta_1(t)$ and $\theta_2(t)$ be two trajectories of the ODE starting at different initial conditions, and let $\tilde{\theta}(t)=\theta_1(t)-\theta_2(t)$. From \eqref{Eq:ODE01}, we get
\begin{displaymath}
\frac{d}{dt}\snorm{\tilde{\theta}}^2_2=-2\tilde{\theta}^\top\Sigma_\pi\tilde{\theta}+2\gamma\EE[\pi]{\big(\phi_x\tilde{\theta}\big)\big(\phi^{\theta_1}_y\theta_1-\phi^{\theta_2}_y\theta_2\big)}.
\end{displaymath}

Notice now that, from the definition of $\phi^{\theta_1}_y$ and $\phi^{\theta_2}_y$, 
\begin{align*} 
\phi^{\theta_1}_y\theta_2&\leq\phi^{\theta_2}_y\theta_2 & \phi^{\theta_2}_y\theta_1&\leq\phi^{\theta_1}_y\theta_1.
\end{align*} 
Taking this into account and defining the sets $S_+=\sset{(x,a)\mid\phi^\top(x,a)\tilde{\theta}>0}$ and $S_-=\X\times\A-S_+$, the previous expression becomes
\begin{align*}
\frac{d}{dt}\snorm{\tilde{\theta}}^2_2\leq-2\tilde{\theta}^\top\Sigma_\pi\tilde{\theta}&+2\gamma\EE[\pi]{\big(\phi_x\tilde{\theta}\big)\big(\phi^{\theta_1}_y\tilde{\theta}\big)\I_{S_+}}\\
  &+2\gamma\EE[\pi]{\big(\phi_x\tilde{\theta}\big)\big(\phi^{\theta_2}_y\tilde{\theta}\big)\I_{S_-}},
\end{align*}
where $\I_S$ represents the indicator function for the set $S$. Applying H\"{o}lder's inequality to each of the expectations above, we get
\begin{align*}
\lefteqn{\frac{d}{dt}\snorm{\tilde{\theta}}^2_2\leq-2\tilde{\theta}^\top\Sigma_\pi\tilde{\theta}}\\
 &\quad+2\gamma\sqrt{\EE[\pi]{(\phi_x\tilde{\theta})^2\I_{S_+}}\EE[\pi]{(\phi^{\theta_1}_y\tilde{\theta})^2\I_{S_+}}}\\
 &\quad+2\gamma\sqrt{\EE[\pi]{(\phi_x\tilde{\theta})^2\I_{S_-}}\EE[\pi]{(\phi^{\theta_2}_y\tilde{\theta})^2\I_{S_-}}}
\end{align*}
A few simple computations thus finally yield
\begin{align*}
\frac{d}{dt}\snorm{\tilde{\theta}}^2_2
  &\leq-2\tilde{\theta}^\top\Sigma_\pi\tilde{\theta}\\
  &+2\gamma\sqrt{2\tilde{\theta}^\top\Sigma_\pi\tilde{\theta}\max\big(\tilde{\theta}^\top\Sigma^*_\pi(\theta_1)\tilde{\theta},\tilde{\theta}^\top\Sigma^*_\pi(\theta_2)\tilde{\theta}\big)}.
\end{align*}
Since, by assumption, $2(1+\eta)\gamma^2\tilde{\theta}^\top\Sigma^*_\pi(\theta_i)\tilde{\theta}\leq\tilde{\theta}^\top\Sigma_\pi\tilde{\theta}$, $i=1,2$, we can conclude from the expression above that
\begin{align*}
\frac{d}{dt}\snorm{\tilde{\theta}}^2_2
  &\leq-2\Big(1-\frac{1}{\sqrt{1+\eta}}\Big)\tilde{\theta}^\top\Sigma_\pi\tilde{\theta}\leq-c\snorm{\tilde{\theta}}^2_2,
\end{align*}
with $c=2(1-1/\sqrt{1+\eta})\lambda_{\min}(\Sigma_\pi)>0$ and $\lambda_{\min}(\Sigma_\pi)$ is the smallest eigenvalue of $\Sigma_\pi$. This means that any two trajectories of the ODE \eqref{Eq:ODE01} approach each other exponentially fast, \ie the ODE is contracting. In particular, the trajectories of the ODE are bounded and, the ODE being autonomous (\ie time-invariant), it has a unique equilibrium point, which is globally asymptotically stable \citep[see, e.g.,][]{lohmiller98automatica}, and verifies the recursive relation
\begin{equation}\label{Eq:EqPoint}
\theta^*=\Sigma_\pi^{-1}\EE[\pi]{\phi_x^\top\big(r(x,a,y)+\gamma\phi^{\theta^*}_y\theta^*\big)}.
\end{equation}
Since $\Sigma_\pi$ is, by assumption, positive definite, the inverse in \eqref{Eq:EqPoint} is well-defined. Multiplying \eqref{Eq:EqPoint} by $\phi^\top(x,a)$ on both sides yields the desired result.
\end{proof}

It is now important to observe that condition \eqref{Eq:Cond01} is quite restrictive. For example, if the features have the same norm for all actions, then $\Sigma_\pi$ and $\Sigma^*_\pi(\theta)$ have the same trace, and \eqref{Eq:Cond01} can only hold if $\gamma<1/\sqrt{2}$. More generally, condition \eqref{Eq:Cond01} requires the second moments of the features of the greedy actions, for every $\theta$, to be dominated by those of the actions selected by the learning policy, with a margin that grows as $\gamma$ approaches $1$. Such condition will seldom be met in practice. 

On the other hand, from a more qualitative perspective, the most interesting aspect of the above result is that it explicitly relates how much information the learning policy provides about greedy policies, as a function of $\gamma$. To better understand this, notice that each policy $\pi$ is associated with a particular invariant measure on the induced chain $(\X,\P_\pi)$. In particular, the measure associated with the learning policy may be very different from the one induced by the greedy/optimal policy. Taking into account the fact that $\gamma$ measures, in a sense, the ``importance of the future'', Theorem~\ref{Theo:QLFAConv} suggests, intuitively, that:

\emph{In problems where the performance of the agent greatly depends on future rewards $(\gamma\approx1)$, the information provided by the learning policy can only be ``safely generalized'' to nearby greedy policies, in the sense of \eqref{Eq:Cond01}. In problems where the performance of the agent is less dependent on future rewards $(\gamma\ll1)$, the information provided by the learning policy can be safely generalized to more general greedy policies.} 

Suppose then that the maximization in the update equation \eqref{Eq:QLFA} is to be replaced by a local maximization. In other words, instead of maximizing over all actions in $\A$, the algorithm should maximize over a small neighborhood of the learning policy $\pi$ (in policy space). The difficulty with this approach is that such maximization can be hard to implement. \emph{Importance sampling} can be used to estimate, from samples obtained with the learning policy $\pi$, the continuation values of a target policy $\pi_\theta$ chosen in such neighborhood, although it does not, by itself, select such policy. The algorithm thus obtained, which resembles in many aspects the one proposed in \citep{Precup01icml}, is described by the update rule
\begin{equation}\label{Eq:modQLFA}
\theta_{t+1}=\theta_t+\alpha_t\phi(x_t,a_t)\Hat{\Delta}_t,
\end{equation}
where the modified temporal difference $\Hat{\Delta}_t$ is given by
\begin{equation}\label{Eq:modTD02}
\Hat{\Delta}_t=r_t+\gamma\frac{\pi_\theta(x_{t+1},a_{t+1})}{\pi(x_{t+1},a_{t+1})}Q_{\theta_t}(x_{t+1},a_{t+1})-Q_{\theta_t}(x_t,a_t),
\end{equation}
where $\pi_\theta$ is, for example, a $\theta$-dependent $\eps$-greedy policy close to the learning policy $\pi$.%
\footnote{Given $\eps>0$, a policy $\pi$ is \emph{$\eps$-greedy} with respect to a function $Q_\theta\in\Q$ if, at each state $x\in\X$, it chooses a random action with probability $\eps$ and a greedy action $a\in\A^\theta_x$ with probability $(1-\eps)$.}
We do not establish the convergence of this variant here. A possible implementation of such algorithm is sketched in Algorithm~\ref{Alg:modQLFA}. We denoted by $\pi_N$ the behavior policy at iteration $N$ of the algorithm; in the stopping condition for the algorithm, any adequate policy norm can be used.

\begin{algorithm}[!htb]
\caption{Modified $Q$-learning.}\label{Alg:modQLFA}
  \begin{algorithmic}[1]
    \REQUIRE Initial policy $\pi_0$, tolerance $\delta>0$;
    \STATE Initialize $\theta_0$, $X_0=x_0$, sample $A_0\sim\pi_0(x_0,\cdot)$, and set $N=0$;
    \FOR{$t=0$ until $T-1$}%
      \label{Alg:Step2}
    	\STATE Sample next-state $X_{t+1}\sim\P_{a_t}(x_t,\cdot)$;
    	\STATE $r_t=r(x_t,a_t,x_{t+1})$;
    	\STATE Sample $A_{t+1}\sim\pi_N(x_{t+1},\cdot)$;
    	\STATE Update $\theta_t$ according to \eqref{Eq:modQLFA};
    \ENDFOR
    \STATE Set $\pi_{N+1}(x,a)=\pi_{\theta_T}(x,a)$;
  	\STATE $N=N+1$;
  	\IF{$\norm{\pi_{N}-\pi_{N-1}}<\delta$}
  		\RETURN $\pi_N$;
  	\ELSE
  		\STATE Set $\theta_0=\theta_T$ and $x_0=x_T$, sample $A_0\sim\pi_N(x_0,\cdot)$, and goto \ref{Alg:Step2};
  	\ENDIF
  \end{algorithmic}
\end{algorithm}

\subsection{{\SARSA} with Linear Function Approximation}%
\label{Subsec:SARSA}

The analysis in the previous subsection suggests that on-policy algorithms may potentially yield more reliable convergence properties. Such fact has already been observed in \citep{Vanroy96-b,Perkins02icml}. In this subsection we thus focus on on-policy algorithms. We analyze the convergence of {\SARSA} when combined with linear function approximation. In our main result, we recover the essence of the result in \citep{Perkins03nips}, although in a somewhat different setting. The main differences between our work and that in \citep{Perkins03nips} are discussed further ahead.

Once again, we consider a family $\Q$ of real-valued functions, the linear span of some fixed set of $M$ functions $\phi_i:\X\times\A\to\R$, and derive an on-policy algorithm to compute a parameter vector $\theta^*$ such that $\phi^\top(x,a)\theta^*$ approximates the optimal $Q$-function. To this purpose, and unlike what has been done so far, we now consider a \emph{$\theta$-dependent} learning policy $\pi_\theta$ verifying $\pi_\theta(x,a)>0$ for all $\theta$. In particular, we consider at each time step a learning policy $\pi_{\theta_t}$ that verifies $\pi_{\theta_t}(x,a)\geq\eps/\abs{\A}$ (for example, a mixture of uniform exploration, with probability $\eps$, and a softmax policy with respect to $\phi^\top(x,a)\theta_t$) and is Lipschitz continuous with respect to $\theta$, with Lipschitz constant $C$ (with respect to some preferred norm). We further assume that, for every fixed $\theta$, the Markov chain $(\X,\P_\theta)$ induced by such policy is uniformly ergodic. We assume the functions $\phi_i,i=1,\ldots,M$, are bounded and linearly independent in $L^2(\mu_{\pi_\theta})$, where $\mu_{\pi_\theta}$ is the state-action measure defined in Section~\ref{Subsec:Thms} for the policy $\pi_\theta$.

Let then $\set{x_t}$, $\set{a_t}$ and $\set{r_t}$ be sampled trajectories of states, actions and rewards obtained from the MDP $\M=\MDP$ using (at each time-step) the $\theta$-dependent policy $\pi_{\theta_t}$. The update rule for our approximate {\SARSA} algorithm is:
\begin{equation}\label{Eq:SARSA}
    \theta_{t+1}=\theta_t+\alpha_t\phi(x_t,a_t)\Delta_t,
\end{equation}
where $\Delta_t$ is the temporal difference at time $t$,
\begin{displaymath}
\Delta_t=r_t+\gamma\phi^\top(x_{t+1},a_{t+1})\theta_t-\phi^\top(x_t,a_t)\theta_t.
\end{displaymath}
In order to use the {\SARSA} algorithm above to approximate the optimal $Q$-function, it is necessary to slowly decay the exploration to zero (for the example above, both $\eps$ and the temperature of the softmax policy), while guaranteeing the learning policy to verify the necessary regularity conditions (namely, Lipschitz continuous w.r.t. $\theta$). However, as will soon become apparent, decreasing the exploration rate to zero will render our convergent result (and other related results) not applicable.

We are now in position to introduce our main result.

\begin{theorem}\label{Theo:SARSAConv}
Let $\M$, $\pi_{\theta_t}$ and $\set{\phi_i,i=1,\ldots,M}$ be as defined above, with $\X$ finite. Let $C$ be the Lipschitz constant of the learning policy $\pi_\theta$ with respect to $\theta$. Assume that the step-size sequence verifies
\begin{align*}
  \sum_t\alpha_t&=\infty & \sum_t\alpha_t^2&<\infty.
\end{align*}
Then, there is $C_0>0$ such that, if $C<C_0$, the algorithm in \eqref{Eq:SARSA} converges w.p.1.
\end{theorem}
\begin{proof}
We again use an ODE argument to establish the statement of the theorem. Since $\X$ is finite, the auxiliary results in Appendix~\ref{App:Aux} hold.

As before, the assumptions on $(\X,\P_\theta)$ and basis functions $\set{\phi_i,i=1,\ldots,M}$ and the fact that the learning policy is Lipschitz continuous with respect to $\theta$ and verifies $\pi(x,a)>0$ ensure the applicability of Theorem~17 in page~239 of \citep{Benveniste90}. Therefore, the convergence of the algorithm can be analyzed in terms of the stability of the associated ODE:
\begin{equation}\label{Eq:ODE02}
  \dot{\theta}=\E_{\theta}\Big[\phi_x^\top\big(r(x,a,y)+\gamma\phi_y\theta-\phi_x\theta\big)\Big].
\end{equation}
Notice that the expectation is taken with respect to the invariant measure of the chain and learning policy, both $\theta$-dependent. To establish global asymptotic stability, we re-write \eqref{Eq:ODE02} as
\begin{displaymath}
\dot{\theta}(t)=\mathbf{A}_\theta\theta(t)+\mathbf{b}_\theta
\end{displaymath}
where
\begin{align*}
  \mathbf{A}_\theta&=\EE[\theta]{\phi_x^\top\big(\gamma\phi_y-\phi_x\big)};&
  \mathbf{b}_\theta&=\EE[\theta]{\phi_x^\top{r}(x,a,y)}.
\end{align*}
An equilibrium point of \eqref{Eq:ODE02} must verify \mbox{$\mathbf{A}_{\theta^*}\theta^*+\mathbf{b}_{\theta^*}=0$}, and the existence of such equilibrium point, with $\snorm{\theta^*}_2\leq\rho$ for some constant $\rho>0$ that does not depend on $C$, is established in Theorem~\ref{Theo:Fixed-point} in the appendix \citep[see also][Theorem~5.1]{farias00jota}. Let $\tilde{\theta}(t)=\theta(t)-\theta^*$. Then,
\begin{align*}
\frac{d}{dt}\snorm{\tilde{\theta}}_2^2&=2\tilde{\theta}^\top\big(\mathbf{A}_\theta\theta+\mathbf{b}_\theta\big)=\\
  &=2\tilde{\theta}^\top\big(\mathbf{A}_\theta\theta-\mathbf{A}_{\theta^*}\theta^*+\mathbf{b}_\theta-\mathbf{b}_{\theta^*}\big).
\end{align*}
The previous expression thus becomes
\begin{align*}
\lefteqn{\frac{d}{dt}\snorm{\tilde{\theta}}_2^2=2\tilde{\theta}^\top\mathbf{A}_\theta\tilde{\theta}}\\
 &+2\tilde{\theta}^\top(\mathbf{A}_\theta-\mathbf{A}_{\theta^*})\theta^*+2\tilde{\theta}^\top(\mathbf{b}_\theta-\mathbf{b}_{\theta^*}).
\end{align*}
By Lemma~\ref{Lemma:Negative-definite-A} in the appendix, there is $c>0$, which does not depend on $C$, such that $\tilde{\theta}^\top\mathbf{A}_\theta\tilde{\theta}\leq-c\snorm{\tilde{\theta}}_2^2$ for all $\theta$.
The fact that the learning policy is assumed Lipschitz w.r.t. $\theta$ and the uniform ergodicity of the corresponding induced chain implies that $\mathbf{A}_\theta$ and $\mathbf{b}_\theta$ are also Lipschitz w.r.t. $\theta$, with constants $c_A$ and $c_b$ proportional to $C$ (Lemma~\ref{Lemma:Lipschitz-A,b} in the appendix). Letting $\lambda=c_A\rho+c_b$, the above expression can be written as
\begin{displaymath}
\frac{d}{dt}\snorm{\tilde{\theta}}_2^2\leq-2(c-\lambda)\snorm{\tilde{\theta}}_2^2.
\end{displaymath}
Since $c_A$ and $c_b$ are proportional to $C$, $\lambda$ goes to zero with $C$ and, therefore, for $C$ sufficiently small, $c-\lambda>0$.
Therefore, $\theta^*$ is the unique equilibrium point of the ODE \eqref{Eq:ODE02}, which is globally asymptotically stable, and the conclusion of the theorem follows.
\end{proof}

Several remarks are now in order. First of all, Theorem~\ref{Theo:SARSAConv} basically states that, for fixed $\eps$, if the dependence of the learning policy $\pi_\theta$ can be made sufficiently ``smooth'', then {\SARSA} converges w.p.1. This result is similar to the result in \citep{Perkins03nips}, although the algorithms are not exactly similar: we consider a continuing task, while the algorithm featured in \citep{Perkins03nips} is implemented in an episodic fashion. Furthermore, in our case, convergence was established using an ODE argument, instead of the contraction argument in \citep{Perkins03nips}. Nevertheless, both methods of proof are, in its essence, equivalent and the results in both papers concordant. 

A second remark is related with the implementation of {\SARSA} with a decaying exploration policy. The analysis of one such algorithm could be conducted using, once again, an ODE argument. In particular, {\SARSA} could be described as a two-time-scale algorithm: the iterations of the main algorithm (corresponding to \eqref{Eq:SARSA}) would develop on a faster time-scale and the decaying exploration rate would develop at a slower time-scale. The analysis in \citep{borkar97scl} could then be replicated. However, as the exploration decays to zero, the learning policy approaches the greedy policy w.r.t. $\theta$ which is, in general, discontinuous. Therefore, there is little hope that the smoothness condition in Theorem~\ref{Theo:SARSAConv} (or its equivalent in \citep{Perkins03nips}) can be met as the exploration approaches zero. Moreover, as $\eps$ approaches zero, the uniform lower bound on the learning policies used in Appendix~\ref{App:Aux} is lost.

\section{Discussion}%
\label{Sec:Conc}

We now briefly discuss some of the assumptions in the above theorems. 

We start by emphasizing that all stated conditions are only \emph{sufficient}, meaning that it is possible that convergence may occur even if some (or all) fail to hold. We also discuss the relation between our results and other related works from the literature.

Secondly, uniform ergodicity of a Markov chain essentially means that the chain quickly converges to a stationary behavior uniformly over the state-space and we can study any properties of the stationary chain by direct sampling.%
\footnote{Explicit bounds on the rate of convergence to stationarity are available in the literature \citep{Meyn94,diaconis96aap,Rosenthal02}. However, for general chains, such bounds tend to be loose.}
This property and the requirement that $\pi(x,a)>0$ for all $a\in\A$ and $\mu_X$-almost all $x\in\X$ can be interpreted as a continuous counterpart to the usual condition that all state-action pairs are visited infinitely often. In fact, uniform ergodicity implies that all the regions of the state-space with positive $\mu_X$ measure are ``sufficiently'' visited \citep{Meyn93}, and the condition $\pi(x,a)>0$ ensures that, at each state, every action is ``sufficiently'' tried. It appears to be a standard requirement in this continuous scenario, as it has also been used in other works \citep{Vanroy96-b,Singh94nips,Perkins03nips}.\footnote{Most of these works make use of geometric ergodicity which, since we admit a compact state-space, is a consequence of uniform ergodicity.} The requirement that $\pi(x,a)>0$ for all $a\in\A$ and $\mu_X$-almost all $x\in\X$ also corresponds to the concept of \emph{fair control} as introduced in \citep{borkar00peis}. 

We also remark that the divergence example in \citep{gordon96tr} is due to the fact that the learning policy fails to verify the Lipschitz continuity condition stated in Theorem~\ref{Theo:SARSAConv} (as discussed in \citep{Perkins02icml}).

\subsection{Related Work}

In this paper, we analyzed how RL algorithms can be combined with linear function approximation to approximate the optimal $Q$-function in MDPs with infinite state-spaces. In the last decade or so, several authors have addressed this same problem from different perspectives. We now briefly discuss several such approaches and their relation to the results in this paper.

One possible approach is to rely on \emph{soft-state aggregation} \citep{Singh94nips,gordon95tr,Vanroy96-a}, partitioning the state-space into ``soft'' regions. Treating the soft-regions as ``hyper-states'', these methods then use standard learning methods (such as $Q$-learning or {\SARSA}) to approximate the optimal $Q$-function. The main differences between such methods and those using linear function approximation (such as the ones portrayed here) are that, in the former, only one component of the parameter vector is updated at each iteration and the basis functions are, by construction, restrained to be positive and to add to one at each point of the state-space.

Sample-based methods \citep{Ormoneit02,Szepesvari04icml} further generalize the applicability of soft-state aggregation methods by using \emph{spreading functions/kernels} \citep{Ribeiro96icics}. Sample-based methods thus exhibit superior convergence rate when compared with simple soft-state aggregation methods, although under somewhat more restrictive conditions.

Finally, RL with general linear function approximation was thoroughly studied in \citep{Vanroy96-b,Tadic01}. Posterior works extended the applicability of such results. In \citep{Precup01icml}, an off-policy convergent algorithm was proposed that uses an importance-sampling principle similar to the one described in Section~\ref{Sec:RLFA}. In \citep{Perkins03nips}, the authors establish the convergence of {\SARSA} with linear function approximation. 

\subsection{Concluding Remarks}

We conclude by observing that all methods analyzed here as well as those surveyed above experience a degradation in performance as the distance between the target function and the chosen linear space increases. If the functions in the chosen linear space provide only a poor approximation of the desired function, there are no practical guarantees on the usefulness of such approximation. The error bounds derived in \citep{Vanroy96-b} are reassuring in that they state that the performance of approximate TD ``gracefully'' degrades as the distance between the target function and the chosen linear space increases. Although we have not addressed such topic in our analysis, we expect the error bounds in \citep{Vanroy96-b} to carry with little changes to our setting. Finally, we make no use of eligibility traces in our algorithms. However, it is just expectable that the methods described herein can easily be adapted to accommodate for eligibility traces, this eventually yielding better approximations (with tighter error bounds) \citep{Vanroy96-b}.

\section*{Acknowledgements}

This version of the paper corrects the one that was presented at the 25th International Conference on Machine Learning. We corrected condition~\eqref{Eq:Cond01} of Theorem~\ref{Theo:QLFAConv} and its proof, as well as the importance-sampling update~\eqref{Eq:modTD02}. We also made the assumptions of Theorems~\ref{Theo:QLFAConv} and~\ref{Theo:SARSAConv} more precise and revised the proof of Theorem~\ref{Theo:SARSAConv}, adding an appendix with the auxiliary results it uses.

The authors would like to acknowledge the many useful comments from Doina Precup and the unknown reviewers. Francisco S. Melo acknowledges the support from the ICTI and the Portuguese FCT, under the Carnegie Mellon-Portugal Program. Sean Meyn gratefully acknowledges the financial support from the National Science Foundation (ECS-0523620). Isabel Ribeiro was supported by the FCT in the frame of ISR-Lisboa pluriannual funding. Any opinions, findings, and conclusions or recommendations expressed in this material are those of the authors and do not necessarily reflect the views of the NSF.

\bibliographystyle{mlapa}
\bibliography{biblio}

\appendix

\section{Auxiliary Results}%
\label{App:Aux}

This appendix establishes the auxiliary results used in the proof of Theorem~\ref{Theo:SARSAConv}. Throughout the appendix, the state space $\X$ is finite and the assumptions of Theorem~\ref{Theo:SARSAConv} hold. We denote by $\mu_\theta$ the invariant distribution of the chain $(\X,\P_\theta)$, so that the expectation $\E_\theta$ in \eqref{Eq:ODE02} is taken with $x$ distributed according to $\mu_\theta$, $a$ distributed according to $\pi_\theta(x,\cdot)$, $y$ distributed according to $\P_a(x,\cdot)$ and, in $\phi_y$, the action at $y$ distributed according to $\pi_\theta(y,\cdot)$. Finally, $\norm{\cdot}_2$ denotes the Euclidean norm of a vector and the corresponding induced norm of a matrix.

By assumption, the learning policies all belong to the set of policies verifying
\begin{displaymath}
\pi(x,a)\geq\frac{\eps}{\abs{\A}}
\end{displaymath}
for all $x$ and $a$. This set is compact. Moreover, all policies in this set assign positive probability to every action and, therefore, induce chains with the same transitions with positive probability. Since the chains induced by the learning policies are uniformly ergodic, so are the chains induced by all policies in the set. By assumption, the features are linearly independent in $L^2(\mu_{\pi_\theta})$ for every $\theta$. Since all policies in the set induce chains with the same recurrent states, the measures $\mu_\pi$ of all policies $\pi$ in the set have the same support. Hence, the features are linearly independent in $L^2(\mu_\pi)$ and the matrix $\Sigma_\pi$ is positive definite for every policy $\pi$ in the set. Several constants below are obtained as maxima or minima over this set of continuous functions of the policy, and are thus finite (or positive). Such constants depend on the MDP, the features, $\gamma$ and $\eps$, but not on $C$.

\begin{lemma}\label{Lemma:Lipschitz-mu}
There is a constant $c_\mu>0$, proportional to $C$, such that, for all $\theta_1$, $\theta_2$ and all $x\in\X$,
\begin{displaymath}
\abs{\mu_{\theta_1}(x)-\mu_{\theta_2}(x)}\leq c_\mu\norm{\theta_1-\theta_2}_2.
\end{displaymath}
\end{lemma}
\begin{proof}
The transition probabilities of the chain $(\X,\P_\theta)$ are
\begin{displaymath}
\P_\theta(x,y)=\sum_{a\in\A}\pi_\theta(x,a)\P_a(x,y).
\end{displaymath}
Since the learning policy is Lipschitz continuous with constant $C$, there is a constant $c_0$, which depends only on $\abs{\A}$ and on the norm used in Section~\ref{Subsec:SARSA}, such that
\begin{displaymath}
\abs{\P_{\theta_1}(x,y)-\P_{\theta_2}(x,y)}\leq c_0C\norm{\theta_1-\theta_2}_2.
\end{displaymath}
The invariant distribution of an ergodic chain varies continuously with its transition probabilities. More precisely, the variation in the invariant distribution is bounded by the variation in the transition probabilities, multiplied by a condition number that depends continuously on the chain \citep{Seneta93spl}. The conclusion follows by taking $c_\mu$ as $c_0C$ times the maximum of this condition number over the chains induced by the set of policies above.
\end{proof}

\begin{lemma}\label{Lemma:Lipschitz-A,b}
There are constants $c_A,c_b>0$, proportional to $C$, such that, for all $\theta_1$ and $\theta_2$,
\begin{align*}
\norm{\mathbf{A}_{\theta_1}-\mathbf{A}_{\theta_2}}_2&\leq c_A\norm{\theta_1-\theta_2}_2,\\
\norm{\mathbf{b}_{\theta_1}-\mathbf{b}_{\theta_2}}_2&\leq c_b\norm{\theta_1-\theta_2}_2.
\end{align*}
\end{lemma}
\begin{proof}
Writing the expectation $\E_\theta$ explicitly, we have
\begin{multline*}
\mathbf{A}_\theta=\sum_{x,a}\mu_\theta(x)\pi_\theta(x,a)\phi_x^\top\\
\cdot\Big(\gamma\sum_{y,b}\P_a(x,y)\pi_\theta(y,b)\phi^\top(y,b)-\phi_x\Big)
\end{multline*}
and
\begin{displaymath}
\mathbf{b}_\theta=\sum_{x,a}\mu_\theta(x)\pi_\theta(x,a)\phi_x^\top\sum_{y}r(x,a,y)\P_a(x,y).
\end{displaymath}
Both are finite sums of products of bounded terms, in which only $\mu_\theta$ and $\pi_\theta$ depend on $\theta$. Since the learning policy is Lipschitz continuous with constant $C$ and, by Lemma~\ref{Lemma:Lipschitz-mu}, $\mu_\theta$ is Lipschitz continuous with constant $c_\mu$, the conclusion follows.
\end{proof}

The next result is essentially due to \citet{Vanroy96-b}.

\begin{lemma}\label{Lemma:Negative-definite-A}
There is a constant $c>0$, which does not depend on $C$, such that, for all $\theta$ and all $v\in\R^M$,
\begin{displaymath}
v^\top\mathbf{A}_\theta v\leq-c\norm{v}_2^2.
\end{displaymath}
\end{lemma}
\begin{proof}
From the definition of $\mathbf{A}_\theta$,
\begin{displaymath}
v^\top\mathbf{A}_\theta v=\gamma\EE[\theta]{(\phi_xv)(\phi_yv)}-\EE[\theta]{(\phi_xv)^2}.
\end{displaymath}
Since $\mu_\theta$ is invariant, the state-action pair at $y$ has the same distribution as $(x,a)$, so that
\begin{displaymath}
\EE[\theta]{(\phi_yv)^2}=\EE[\theta]{(\phi_xv)^2}.
\end{displaymath}
By the Cauchy-Schwarz inequality, we then get
\begin{displaymath}
v^\top\mathbf{A}_\theta v\leq-(1-\gamma)\EE[\theta]{(\phi_xv)^2}=-(1-\gamma)v^\top\Sigma_{\pi_\theta}v,
\end{displaymath}
where $\Sigma_{\pi_\theta}$ is the matrix defined in Section~\ref{Subsec:Thms} for the policy $\pi_\theta$. By the assumption above, the smallest eigenvalue of $\Sigma_\pi$ is positive for every policy in the set, and depends continuously on $\pi$. The conclusion follows by taking $c$ as $1-\gamma$ times the minimum of this eigenvalue over the set.
\end{proof}

\begin{theorem}\label{Theo:Fixed-point}
The equation
\begin{displaymath}
\mathbf{A}_\theta\theta+\mathbf{b}_\theta=0
\end{displaymath}
has a solution $\theta^*$ with $\norm{\theta^*}_2\leq\rho$, for some constant $\rho>0$ that does not depend on $C$.
\end{theorem}
\begin{proof}
By Lemma~\ref{Lemma:Negative-definite-A} and the Cauchy-Schwarz inequality, for all $\theta$ and $v$,
\begin{displaymath}
\norm{\mathbf{A}_\theta v}_2\norm{v}_2\geq-v^\top\mathbf{A}_\theta v\geq c\norm{v}_2^2.
\end{displaymath}
Hence, each matrix $\mathbf{A}_\theta$ is non-singular, with $\norm{\mathbf{A}_\theta^{-1}}_2\leq1/c$. Consider now the map
\begin{displaymath}
\theta\mapsto-\mathbf{A}_\theta^{-1}\mathbf{b}_\theta.
\end{displaymath}
A fixed point of this map is a solution of $\mathbf{A}_\theta\theta+\mathbf{b}_\theta=0$. By Lemma~\ref{Lemma:Lipschitz-A,b} and the uniform bound on the inverses, the map is continuous. Moreover, since the features and the reward are bounded, the norms of the vectors $\mathbf{b}_\theta$ are bounded by a constant that does not depend on $C$. Letting $\rho$ be this constant divided by $c$, the map takes values in the closed ball of radius $\rho$ centered at the origin and, in particular, maps this ball into itself. By Brouwer's fixed-point theorem, it has a fixed point in the ball.
\end{proof}

\end{document}